\documentclass{article}

    \usepackage[preprint]{neurips_data_2024}

\usepackage[utf8]{inputenc} 
\usepackage[T1]{fontenc}    
\usepackage{hyperref}       
\usepackage{url}            
\usepackage{booktabs}       
\usepackage{amsfonts}       
\usepackage{nicefrac}       
\usepackage{microtype}      
\usepackage{xcolor}         

\usepackage{graphicx} 
\usepackage{natbib}
\usepackage{colortbl}
\setcitestyle{square,comma,numbers,sort&compress}
\usepackage{subcaption}
\usepackage{animate}
\usepackage{multirow,multicol}

\def\DatasetName{\emph{MiraData}}
\def\BenchName{\emph{MiraBench}}
\def\ModelName{\emph{MiraDiT}}

\definecolor{Red}{RGB}{192, 0, 0}
\definecolor{Blue}{RGB}{12, 114, 186}
\definecolor{Yellow}{RGB}{218, 169, 20}

\definecolor{HighlightBlue}{RGB}{0, 100, 148}
\definecolor{HighlightRed}{RGB}{230, 57, 70}

\definecolor{LightRed}{HTML}{ffe0e0}
\definecolor{LightBlue}{HTML}{def5ff}
\definecolor{LightYellow}{HTML}{FFF6DB}
\definecolor{LightGreen}{HTML}{eff9f0}

\definecolor{Gray}{HTML}{DCDCDC}

\usepackage{wrapfig,lipsum}

\title{MiraData: A Large-Scale Video Dataset with \\ Long Durations and Structured Captions}

\author{%
\textbf{Xuan Ju$^{1,2}$\thanks{Equal contribution. $^{\dag}$ Project Lead. $^{1}$ARC Lab, Tencent PCG. $^{2}$The Chinese University of Hong Kong.}~~, Yiming Gao$^{1}$\footnote[1]{}~~, Zhaoyang Zhang$^{1\dag}$\footnote[1]{} ~~, Ziyang Yuan$^{1}$, Xintao Wang$^{1}$~~,}\\ 
\textbf{Ailing Zeng$^{2}$, Yu Xiong$^{2}$, Qiang Xu$^{2}$, Ying Shan$^{1}$} \\
\url{https://github.com/mira-space/MiraData}
}

\begin{document}

\maketitle

\begin{center}
\vspace{-0.55cm}
    \captionsetup{type=figure}
   \includegraphics[width=0.95\textwidth,height=0.53\textwidth]{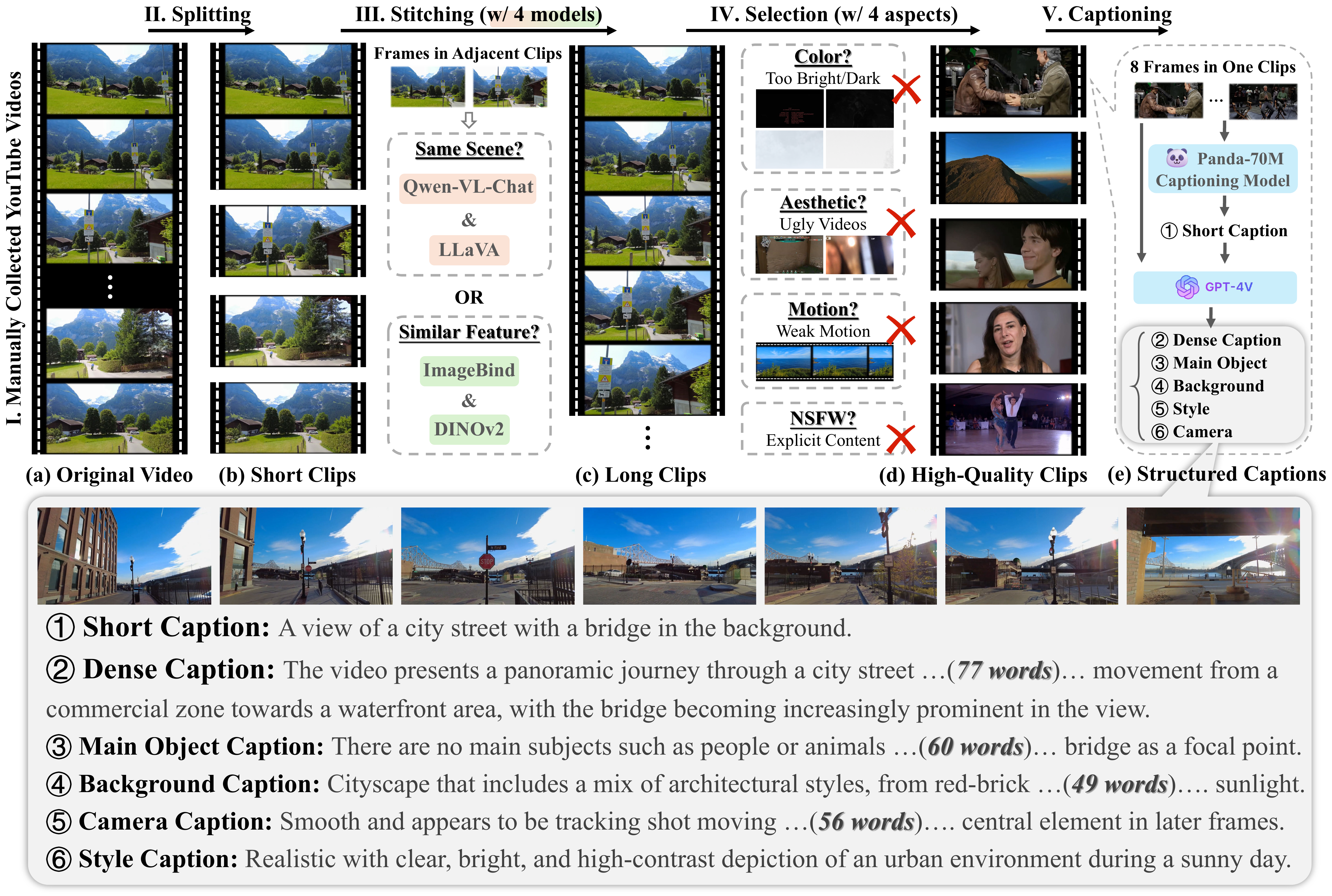}
\vspace{-0.1cm}
    \captionof{figure}{\textbf{Video collection and annotation pipeline.} An example shown at bottom.}
\label{fig:annotation}
\end{center}

\vspace{-0.2cm}

\begin{abstract}
\vspace{-0.3cm}

Sora's high-motion intensity and long consistent videos have significantly impacted the field of video generation, attracting unprecedented attention. However, existing publicly available datasets are inadequate for generating Sora-like videos, as they mainly contain short videos with low motion intensity and brief captions.
To address these issues, we propose \DatasetName, a high-quality video dataset that surpasses previous ones in video duration, caption detail, motion strength, and visual quality. We curate \DatasetName~from diverse, manually selected sources and meticulously process the data to obtain semantically consistent clips. GPT-4V is employed to annotate structured captions, providing detailed descriptions from four different perspectives along with a summarized dense caption.
To better assess temporal consistency and motion intensity in video generation, we introduce \BenchName, which enhances existing benchmarks by adding 3D consistency and tracking-based motion strength metrics. \BenchName~includes 150 evaluation prompts and 17 metrics covering temporal consistency, motion strength, 3D consistency, visual quality, text-video alignment, and distribution similarity.
To demonstrate the utility and effectiveness of \DatasetName, we conduct experiments using our DiT-based video generation model, \ModelName. The experimental results on \BenchName~demonstrate the superiority of \DatasetName, especially in motion strength.

\end{abstract}

\section{Introduction}
\label{sec:introduction}

Recent advances in the Artificial Intelligence and Generative Content (AIGC) field, such as video generation~\cite{sora, svd, snapvideo}, image generation~\cite{pixart,emu, sd3,pixart_sigma}, and natural language processing~\cite{gpt4,llama3}, have been rapidly progressing, thanks to the improvements in data scale and computational power. Previous studies~\cite{pixart,llama3,svd,pixart_sigma} have emphasized that data plays a pivotal role in determining the upper-bound performance of a task. Recently, Sora~\cite{sora}, a text-to-video generation model, shows stunning video generation capabilities far surpassing existing state-of-the-art methods. Sora not only excels in generating high-quality long videos ($10$-$60$ seconds) but also stands out in terms of motion strength, 3D consistency, adherence to real-world physics rules, and accurate interpretation of prompts.  

The first step in constructing Sora-like video generation models is to curate a high-quality dataset since data serves as the foundation. However, existing publicly video datasets, such as WebVid-10M~\cite{webvid10m}, Panda-70M~\cite{panda70m}, and HD-VILA-100M~\cite{hdvila}, primarily consist of short video clips ($5$-$18$ seconds) sourced from unfiltered videos in the internet, which leads to a large proportion of low-quality or low-motion clips and are inadequate for training generating Sora-like models. Moreover, the captions in existing datasets are often short ($12$-$30$ words) and lack the necessary details to describe videos.
These limitations hinder the use of existing datasets for generating long videos with accurate interpretation of prompts. Therefore, there is an urgent need for a comprehensive, high-quality video dataset with long video durations, strong motion strength, and detailed captions.

To tackle these issues, we present \DatasetName, a large-scale, high-quality video dataset featuring long videos (average of $72.1$ seconds) with high motion intensity and detailed structured captions (average of $318$ words). The data curation pipeline is illustrated in Fig.~\ref{fig:annotation}, where we have built an end-to-end pipeline for data downloading, segmentation, filtering, and annotation.
\textbf{I.} To obtain diverse videos, we collect source videos from manually selected channels of various platforms.
\textbf{II \& III.}~We employ multiple models to compare semantic and visual feature information, segmenting videos into long clips with strong semantic consistency by using a mixture of models to detect clips within a video and cut long videos into smaller segments. 
\textbf{IV.} To accommodate high-quality clips, we filter the dataset into five subsets based on aesthetics, motion intensity, and color to select clips with high visual quality and strong motion intensity. 
\textbf{V.} To obtain detailed and accurate descriptions, we first use the state-of-the-art captioner~\cite{panda70m} to generate a short caption and then employ GPT-4V to enrich it, resulting in the dense caption. To provide fine-grained video descriptions across multiple perspectives, we further design structured captions, which include descriptions of the video's main subject, background, camera motion, and style. 
To this end, Statistical results shown in Tab.~\ref{tab:related_work_dataset} and Tab.~\ref{tab:compare_previous_dataset} encompassing video duration, caption length and elaboration, motion strength, and video quality demonstrate \DatasetName's superiority over previous datasets.

To further analyze the gap between generated videos and high-quality videos, we observe that existing benchmarks lack a comprehensive evaluation of 3D consistency and motion intensity in generated videos. To address this issue, we propose  \textbf{\BenchName}, an enhanced benchmark that builds upon existing benchmarks by adding 3D consistency and tracking-based motion strength metrics. Specifically, MiraBench includes 17 metrics that comprehensively cover various aspects of video generation, such as temporal consistency, motion strength, 3D consistency, visual quality, text-video alignment, and distribution similarity. To evaluate the effectiveness of captions, we introduce 150 evaluation prompts in MiraBench, consisting of short captions, dense captions, and structured captions. These prompts provide a diverse set of challenges for assessing the performance of text-to-video generation models. To validate the effectiveness of our \DatasetName~, we conduct experiments using our DiT-based video generation model, \textbf{\ModelName}. Experimental results show the superiority of our model trained on \DatasetName, when compared to the same model trained on WebVid-10M and other state-of-art open-source methods on motion strength, 3D consistency and other metrics in \BenchName.

\section{Related Work}
\label{sec:related_work}

\subsection{Video-Text Datasets}

Large-scale training on image-text pairs~\cite{clip,align,coyo,laion,motionx} has been proven effective in text-to-image generation~\cite{latentdiffusion,imagen,dalle3} and vision-language representation learning~\cite{blip2,seed}, showing emergent ability with model and data scaling-up. 
Recent achievements such as Sora~\cite{sora} suggest that similar capabilities can be observed in the realm of videos, where data availability and computational resources emerge as crucial factors.
However, previous text-video datasets, as shown in Tab.~\ref{tab:related_work_dataset}, are constrained by short durations, limited caption lengths, and poor visual quality.

Considering the domain of general video generation, a significant portion of open-source text-video datasets is unsuitable due to issues such as noisy text labels, low resolution, and limited domain coverage.
Thus the majority of video generation models with impressive performance~\cite{imagenvideo,snapvideo,emuvideo,videodiffusionmodel,walt,pyoco,videopoet} rely heavily on internal datasets for training, which restricts transparency and usability.
The commonly used open-source text-video dataset for video generation~\cite{makeavideo,magicvideo,alignyourlatents,latentshift,dynamicrafter,videocrafter1,videocrafter2,i2vgenxl,videocomposer,videolcm,moonshot} is WebVid-10M~\cite{webvid10m}. However, it contains a prominent watermark on videos, requiring additional fine-tuning on image datasets (e.g., Laion~\cite{laion5b}) or internal high-quality video datasets to remove the watermark. Recently, Panda-70M~\cite{panda70m}, InternVid~\cite{internvid}, and HD-VG-130M~\cite{videofactory} have been proposed and targeted for video generation. Panda-70M and InternVid aim to extract precise textual annotations using multiple caption models, while HD-VG-130M emphasizes the selection of high-quality videos. But none of them systematically considers correct video splitting, visual quality filtering, and accurate textual annotation at all three levels during the data collection process. More importantly, all previous datasets consist of videos with short durations and limited text lengths, which restricts their suitability for long video generation with fine-grained textual control.

\begin{table}[htbp]
    \centering
    \small
    \caption{\textbf{Comparison of \DatasetName~and pervious large-scale video-text datasets.} Datasets are sorted based on average text length. Datasets with \colorbox{Gray}{gray} background are used in a text-to-video generation. \DatasetName~significantly surpasses previous datasets in average text and video length. }
    \label{tab:related_work_dataset}
    \setlength\tabcolsep{4.5pt}
  \begin{tabular}{lccccccccc}
    \toprule
    Dataset & Avg text len  & \multicolumn{2}{c}{Avg / Total video len} &  Year & Text & Domain & Resolution \\
    \midrule
    HowTo100M~\cite{howto100m}  & 4.0 words   & 3.6s  & 135Khr    & 2019 & ASR     & Open  & 240p  \\
    LSMDC~\cite{lsmdc}     & 7.0 words   & 4.8s  & 158h        & 2015 & Manual    & Movie     & 1080p \\
    DiDeMo~\cite{didemo}    & 8.0 words    & 6.9s  & 87h         & 2017 & Manual    & Flickr    & -     \\
    YouCook2~\cite{youcook2}  & 8.8 words & 19.6s & 176h      & 2018 & Manual    & Cooking       & -     \\
    MSR-VTT~\cite{msrvtt}    & 9.3 words    & 15.0s & 40h       & 2016 & Manual    & Open       & 240p  \\
    \rowcolor{Gray} HD-VG-130M~\cite{videofactory}    & $\sim$9.6 words       & $\sim$5.1s  & $\sim$184Khr    & 2024 &  Generated    & Open      & 720p     \\
    \rowcolor{Gray} WebVid-10M~\cite{webvid10m}    & 12.0 words   & 18.0s & 52Kh    & 2021 & Alt-Text    & Open  & 360p     \\
    \rowcolor{Gray} Panda-70M~\cite{panda70m}  & 13.2 words  & 8.5s  & 167Khr   & 2024 & Generated & Open     & 720p \\
    ActivityNet~\cite{activitynet}  & 13.5 words   & 36.0s & 849h   & 2017 & Manual    & Action    & -     \\
    VATEX~\cite{vatex}   & 15.2 words    & $\sim$10s & $\sim$115h        & 2019 & Manual    & Open    & - \\
    \rowcolor{Gray} HD-VILA-100M~\cite{hdvila}  & 17.6 words    & 11.7s & 760.3Khr  & 2022 & ASR   & Open   & 720p  \\
    How2~\cite{how2}    & 20.0 words  & 5.8s & 308h  & 2018 & Manual    & Instruct  & -     \\
    \rowcolor{Gray} InternVid~\cite{internvid} & 32.5 words    & 13.4s & 371.5Khr   & 2023 &  Generated               & Open   & 720p  \\
    \midrule
    \rowcolor{Gray} \textbf{\DatasetName~(Ours)}   & 318.0 words  & 72.1s  & 
    16Khr  & 2024 &  Generated & Open    & 720p \\
    \bottomrule
    \end{tabular}
\end{table}

\subsection{Video Generation}

Video generation is a challenging task that have advanced from early GAN-based models~\cite{vgan,mcvd} to more recent diffusion.
Diffusion-based methods have made significant progress in terms of visual quality and diversity in generated videos while entailing a substantial computational cost~\cite{emuvideo,snapvideo}. 
Consequently, researchers often face a trade-off between the quality of the generated videos and the duration of the videos that can be produced within practical computational constraints.

To ensure visual quality under computational resource constraints, previous diffusion-based video generation methods primarily focus on open-domain text-to-video generation with a \textbf{short duration}. Video Diffusion Models~\cite{videodiffusionmodel} is the first to employ the diffusion model for video generation.  
To generate long videos in the absence of corresponding dataset, Make-A-Video~\cite{makeavideo} and NUWA-XL~\cite{nuwaxl} explore coarse-to-fine video generation but suffer from maintaining temporal continuity and producing strong motion magnitude. 
Apart from these explorations of convolution-based architecture~\cite{makeavideo,magicvideo,alignyourlatents,videodiffusionmodel,imagenvideo,pyoco,emuvideo,latentshift,videofactory,videocomposer,videocrafter1,videocrafter2,dynamicrafter,videolcm,moonshot}, transformer-based methods (\textit{e.g.}, WALT~\cite{walt}, Latte~\cite{latte}, and Snap Video~\cite{snapvideo}) become more prevalent recently, offering a better trade-off between computational complexity and performance, as well as improved scalability.

All previous methods can only generate short video clips (\textit{e.g.}, 2 seconds, 16 frames) with weak motion strength. However, the recent success of Sora~\cite{sora} demonstrates the potential of long video generation with enhanced motion strength and strong 3D consistency. With the belief that data is the key to machine learning, we find that existing datasets' (1) short duration, (2) weak motion strength, and (3) short and inaccurate captions are insufficient for Sora-like video generation model training (as shown in Tab.~\ref{tab:related_work_dataset}). 
To address these limitations and facilitate the development of advanced video generation models, we introduce \DatasetName, the first large-scale video dataset specifically designed for long video generation. \DatasetName~features videos with longer durations and structured captions, providing a rich and diverse resource for training models capable of generating extended video sequences with enhanced motion and coherence.

\section{MiraData Dataset}
\label{sec:miradata_dataset}

\DatasetName~is a large-scale text-video dataset with long duration and structured detailed captions. We show the overview of the collection and annotation pipeline of \DatasetName~in Fig.~\ref{fig:annotation}. The final dataset was obtained through a five-step process, which involved collection (in Sec.~\ref{sec:data_collection}), splitting and stitching (in Sec.~\ref{sec:video_splitting_and_stitching}), selection (in Sec.~\ref{sec:video_selection}), and captioning (in Sec.~\ref{sec:video_captioning}).

\subsection{Data Collection}
\label{sec:data_collection}

The source of videos is crucial in determining the dataset's data distribution. 
In video generation tasks, there are typically four key expectations: (1) diverse content, (2) high visual quality, (3) long duration, and (4) large motion strength. 
Existing text-to-video datasets~\cite{panda70m,hdvila,videofactory} mainly consist of videos from YouTube. 
Although YouTube offers a vast collection of diverse videos, a large proportion of the videos lack the necessary aesthetic quality for video generation needs. 
To address all four aspects simultaneously, we select source videos from YouTube, Videvo, Pixabay, and Pexels~\footnote{YouTube: \url{https://www.youtube.com/}, Videvo: \url{https://pixabay.com/}, Pixabay: \url{https://www.videvo.net/}, Pexels: \url{https://www.pexels.com/}}, ensuring a more comprehensive and suitable data source for video generation tasks.

\textbf{YouTube Videos.} Following previous works~\cite{hdvila,panda70m,videofactory}, we include YouTube as one of the video sources. 
However, prior research mainly focuses on collecting diverse videos that are suitable for understanding tasks while giving limited consideration to the need for generation tasks (\textit{e.g.,} duration, motion strength, and visual quality), which are crucial for learning physical laws and 3D consistency.

\begin{wrapfigure}{r}{0.4\textwidth}
    \vspace{-0.5cm}
\begin{center}
            \includegraphics[width=0.4\textwidth]{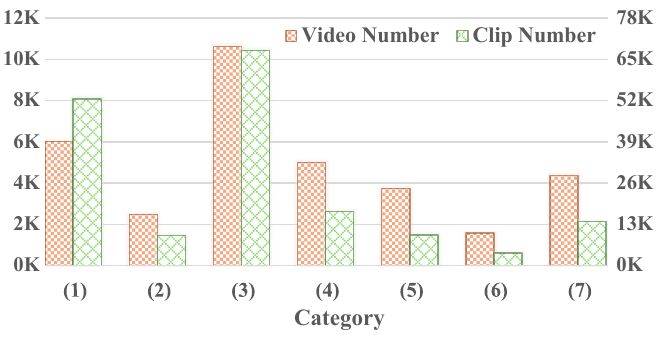}
        \end{center}
    \vspace{-0.4cm}
    \caption{\textbf{The video and video clip distribution of different video categories.} (1) to (7) is explained in Sec.~\ref{sec:data_collection}. }
    \vspace{-0.25cm}
    \label{fig:youtube_crawl_distribution}
\end{wrapfigure}

To address these limitations, we manually select $156$ high-quality YouTube channels that are suitable for generation tasks. 
These channels encompass various categories with rich motion and long video clips, including (1) 3D engine-rendered scenes, (2) city/scenic tours, (3) movies, (4) first-person perspective camera videos, (5) object creation/physical law demonstrations, (6) timelapse videos, and (7) videos showcasing human motion. 
We collect around $68K$ videos with 720p resolution from these YouTube channels ($K$ denotes thousand). 
After the video splitting and stitching operation described in Sec.~\ref{sec:video_splitting_and_stitching}, we obtain around $34K$ videos with $173K$ video clips.
The number of videos and clips for each category are shown in Fig.~\ref{fig:youtube_crawl_distribution}. 
We collect more videos from 3D engine-rendered scenes and movies because they exhibit greater diversity and better visual quality. 
Moreover, the simplicity and consistency of the physical laws in 3D engine-rendered videos are crucial for enabling video generation models to learn and understand physical laws.

Additionally, to ensure data diversity and amount, we also include videos from HD-VILA-100M~\cite{hdvila}. 
Although this dataset contains around 100 million video clips, after the splitting and stitching operation in Sec.~\ref{sec:video_splitting_and_stitching}, only $195K$ clips remain.
This further demonstrates the quality of our selected video sources, as evidenced by a higher retention rate considering video duration and continuity.

\textbf{Videvo, Pixabay, and Pexels Videos.} These three websites offer stock videos and motion graphics free from copyright issues, which are usually exceptionally high-quality videos uploaded by skilled photographers. Although the videos are usually shorter in duration compared to YouTube, they can compensate for the deficiencies in the visual quality of YouTube videos. Therefore, we collect and annotate videos from these websites, which can enhance the generated videos' aesthetics. We finally obtain around $63K$ videos from Videvo, $43K$ videos from Pixabay, and $318K$ videos from Pexels.

\vspace{-0.35cm}

\subsection{Video Splitting and Stitching}
\label{sec:video_splitting_and_stitching}

\begin{figure}[htbp]
\vspace{-0.15cm}
        \begin{center}
            \includegraphics[width=1.\textwidth]{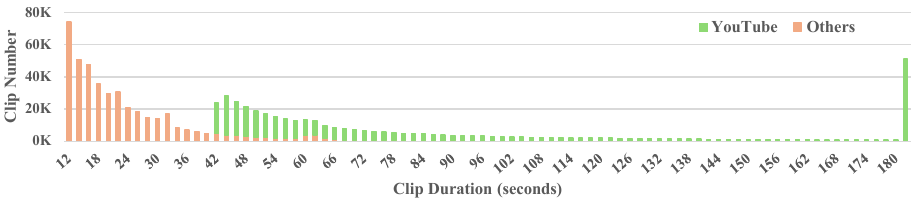}
        \end{center}
\vspace{-0.45cm}
    \caption{\textbf{Distribution of video clip duration from YouTube and other sources}. }
        \label{fig:duration_distribution}
\vspace{-0.45cm}
\end{figure}

An ideal video clip for video generation should have semantically coherent content, either without shot transitions or with strong continuity between transitions. To achieve this, we conduct a two-stage splitting and stitching process on YouTube videos. In the splitting stage, we use shot change detection with a low threshold to divide the video into segments\footnote{We use PySceneDetect content-aware detection with a threshold of $26$}
, ensuring that all distinct clips are extracted. We then stitch short clips together to avoid incorrect separation, considering content-coherent video transitions and accuracy. We employ Qwen-VL-Chat\cite{QwenVL}, LLaVA\cite{llava1,llava2}, ImageBind\cite{imagebind}, and DINOv2\cite{dinov2} to assess whether adjacent short clips should be connected. Vision language models excel in detecting content-coherent transitions, while image feature cosine similarity is more effective in connecting incorrect separations. A connection is made only if both vision language models or both image feature extraction models agree. We retain clips longer than 40 seconds for \DatasetName. Since Videvo, Pixabay, and Pexels videos are naturally in clip form, we select clips longer than 10 seconds to filter for longer videos with greater motion strength. Fig.~\ref{fig:duration_distribution} presents the distribution of video clip duration from YouTube and other sources.

\subsection{Video Selection}
\label{sec:video_selection}

\DatasetName~provides 5 data versions with different quality levels for video generation training, filtered using four criteria: (1) Video Color, (2) Aesthetic Quality, (3) Motion Strength, and (4) Presence of NSFW Content. For Video Color, we filter videos shot in overly bright or dark environments by calculating average color and the color of the brightest and darkest 80\% of frames. Aesthetic Quality is assessed using the Laion-Aesthetic\cite{laion5b} Aesthetic Score Predictor. Motion Strength is measured using the RAFT\cite{raft} algorithm to calculate optical flow between frames. NSFW content is detected using the Stable Diffusion Safety Checker~\cite{latentdiffusion} on 8 evenly selected frames per video. For criteria (1)-(3), we standardize the frame rate to 2 fps and filter videos into four lists based on increasing threshold values. NSFW videos are filtered out from all datasets. The 5 filtered versions contain 788K, 330K, 93K, 42K, and 9K video clips. Details about the filtering process and thresholds are in the supplementary files.

\subsection{Video Captioning}
\label{sec:video_captioning}

As emphasized by PixArt\cite{pixart} and DALL-E 3\cite{dalle3}, the quality and granularity of captions are crucial for text-to-image generation. Given the similarities between image and video generation, detailed and accurate textual descriptions should also play a vital role in the latter. However, previous video-text datasets with meta-information annotations (e.g., WebVid-10M\cite{webvid10m}, HD-VILA-100M\cite{hdvila}) often have incorrect temporal alignment or inaccurate descriptions. Current state-of-the-art video captioning methods generate either simple (e.g., Panda-70M\cite{panda70m}) or inaccurate (e.g., Video-LLaVA\cite{videollava}) captions. To obtain detailed and accurate captions, we use the more powerful GPT-4V~\cite{gpt4v}, which outperforms existing open-source methods.

To enable GPT-4V, a vision language model with image input only, to understand videos, we extract 8 uniformly sampled frames from each video and arrange them in a $2\times4$ grid within a single image. This approach reduces computational cost and facilitates accurate caption generation. Following DALL-E 3\cite{dalle3}, we bias GPT-4V to produce video descriptions useful for learning a text-to-video generation model. We first use Panda-70M\cite{panda70m} to generate a "short caption" describing the main subject and actions, which serves as an additional hint for GPT-4V. The GPT-4V-generated "dense caption" covers the main subject, movements, style, backgrounds, and cameras.

To obtain more detailed, fine-grained, and accurate captions, we propose the use of structured captions. 
In addition to the short and dense captions, structured captions provide further descriptions of crucial elements in the video, including:
(1) Main Object: describes the primary object or subject in the video, capturing their attributes, actions, positions, and movements, 
(2) Background: provides context about the environment or setting, including objects, location, weather, and time, 
(3) Camera Movements: details any camera pans, zooms, or other movements, and 
(4) Video Style: covers the artistic style, as well as the visual and photographic features of the video (\textit{e.g.}, realistic, cyberpunk, and cinematic).

\begin{wrapfigure}{r}{0.5\textwidth}
    \vspace{-0.6cm}
\begin{center}
            \includegraphics[width=0.5\textwidth]{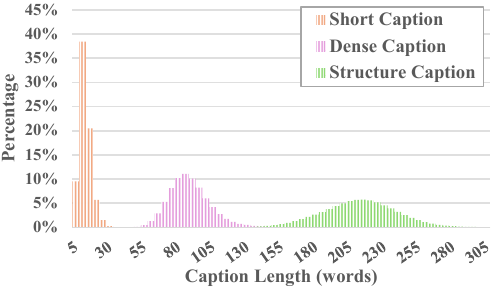}
        \end{center}
    \vspace{-0.4cm}
    \caption{\textbf{Distribution of caption length.} }
    \vspace{-0.4cm}
    \label{fig:word_count_both}
    \vspace{-0.25cm}
\end{wrapfigure}

These structured captions provide extra detailed descriptions from various perspectives, enhancing the richness of the captions. 
With our carefully designed prompt, we can efficiently obtain the video's structured caption from GPT-4V in just one conversation round. 
As demonstrated in Tab.~\ref{tab:related_work_dataset} and Fig.~\ref{fig:word_count_both}, the average caption length of dense descriptions and structured captions has significantly increased to  $90$ and $214$ words respectively, greatly enhancing the descriptive capacity of the captions.

\subsection{Comparison on Numerical Statistics}
\label{sec:comparison_on_numerical_statistics}

We calculate the average frame optical flow strength and aesthetic score on \DatasetName's unfiltered version ($788K$ video clips) and filtered version ($330K$ video clips)  with previous video generation datasets (Panda-70M~\cite{panda70m}, HD-VILA-100M~\cite{hdvila}, InternVid~\cite{internvid}, and WebVid-10M~\cite{webvid10m}). 
For \DatasetName, we calculated the metrics on the full dataset. 
For other datasets, we randomly select $10K$ video clips to save computation costs. 
The frame rate is standardized to 2 for both metrics. 
The results in Tab.~\ref{tab:compare_previous_dataset} show the superiority of \DatasetName, considering both visual quality and motion strength.

\begin{table}[htbp]
 \centering
    \small
        \vspace{-8pt}
    \caption{\textbf{Numerical statics comparison of previous datasets and \DatasetName.}}
    \label{tab:compare_previous_dataset}
    \setlength\tabcolsep{1.3pt}
\begin{tabular}{c|cccc|cc}
\toprule
\textbf{Metrics} & Panda-70M & HD-VILA-100M & InternVid & WebVid-10M & \textbf{MiraData$_{unfilter}$}  & \textbf{MiraData$_{filter}$} \\ \midrule
\textbf{Optical Flow} $\uparrow$    &    4.37       &       4.45       &     3.92      &     1.08       &  \underline{5.22} & \textbf{6.93} \\
\textbf{Aesthetic Score} $\uparrow$     &    4.67       &     4.61         &     4.50      &      4.41     &  \underline{5.01} & \textbf{5.02} \\ \bottomrule
\end{tabular}
    \vspace{-6pt}
\end{table}

\section{MiraBench}
\label{sec:mirabench}

\subsection{Prompt Selection}
\label{sec:prompt_selection}

Following EvalCrafter~\cite{evalcrafter}, we propose four categories: human, animal, object, and landscape. We randomly select 400 video captions, manually curate them for balanced representation across meta-classes, and prioritize captions closely matching the original videos. We select 50 precise video-text pairs, using short, dense, and structured captions as prompts, forming a set of 150 prompts.

\subsection{Metrics Design}
\label{sec:metics_design}

We design $17$ evaluation metrics in \BenchName~from $6$ perspectives, including temporal consistency, temporal motion strength, 3D consistency, visual quality, text-video alignment, and distribution consistency. 
These metrics encompass most of the common evaluation standards used in previous video generation models and text-to-video benchmarks. 
Compared to previous benchmarks like VBench~\cite{vbench}, our metrics place more emphasis on the model's performance with general prompts instead of manually designed prompts and emphasize 3D consistency and motion strength.

\textbf{Temporal Motion Strength.} (1) \textbf{\textit{Dynamic Degree.}} Following previous works~\cite{vbench,internvid}, we use the average distance of optical flow estimated by RAFT~\cite{raft} to estimate the dynamics degree. (2) \textbf{\textit{Tracking Strength.}} In optical flow, the objective is to estimate the velocity of all points within a video frame. This estimation is performed jointly for all points, but the motion is predicted only at an infinitesimal distance. In tracking, the goal is to estimate the motion of points over an extended period. Therefore, the distance of tracking points can better distinguish whether the video involves long-range or minor movements (\textit{e.g.}, camera shake or local movements that move back and forth). As shown in Fig.~\ref{fig:motion_strength} (a), the left figure exhibits a smaller motion distance than the right. However, in Fig.~\ref{fig:motion_strength} (b), the dynamic degree is incorrectly $1.2$ for the left and $0.7$ for the right, suggesting that the left motion is larger. Tracking strength in Fig.~\ref{fig:motion_strength} (c) accurately reflects the moving distance, with $4.1$ for the left and $11.8$ for the right. We use CoTracker~\cite{cotracker} to calculate the tracking path and average the tracking points' distance from the initial frame as the tracking strength metric.

\begin{figure}[htbp]

\centering
    \animategraphics[width=0.95\linewidth, poster=last, autoplay]{30}{videos/tracking_score/frame_}{1}{93}
    \caption{\textbf{Illustration of the difference between tracking strength and optical flow dynamic degree.} \textit{Best viewed with Acrobat Reader. Click the images to play the animation clips.}
    }
    \label{fig:motion_strength}
\end{figure}

\textbf{Temporal Consistency.} (3) \textbf{\textit{DINO (Structural) Temporal Consistency.}} DINO~\cite{dinov2} focuses on structural information. We calculate the cosine similarity of adjacent frames' DINO features to assess structural temporal consistency. (4) \textbf{\textit{CLIP (Semantic) Temporal Consistency.}} We calculate the cosine similarity of adjacent frames' CLIP~\cite{clip} features to assess structural temporal consistency since CLIP focuses on semantic information. (5) \textbf{\textit{Temporal Motion Smoothness.}} Following VBench~\cite{vbench}, we use the motion priors in the video interpolation model AMT~\cite{amt} to calculate the motion smoothness. Since larger motion is expected to contain smaller consistency and vice versa, we multiply \textit{Tracking Strength} by these feature similarities to obtain more reasonable temporal consistency metrics.

\textbf{3D Consistency.} Following GVGC~\cite{gvgc}, we calculate (6) \textbf{\textit{Mean Absolute Error}}, and (7) \textbf{\textit{Root Mean Square Error}} to evaluate video 3D consistency from the perspective of 3D reconstruction.

\textbf{Visual Quality.} (8) \textbf{\textit{Aesthetic Quality}}. We evaluate the
aesthetic score of generated video frames using the LAION aesthetic predictor~\cite{latentdiffusion}. (9) \textbf{\textit{Imaging Quality}}. Following VBench~\cite{vbench}, we evaluate video distortion (\textit{e.g.}, over-exposure, noise, and blur) using the MUSIQ~\cite{musiq} quality predictor.

\textbf{Text-Video Alignment.} We use ViCLIP~\cite{internvid} to evaluate the consistency between video and text. We calculate from $5$ aspects following \BenchName~prompt structure: (10) \textbf{\textit{Camera Alignment}}. (11) \textbf{\textit{Main Object Alignment}}. (12) \textbf{\textit{Background Alignment}}. (13) \textbf{\textit{Style Alignment}}. (14) \textbf{\textit{Overall Alignment}}.

\textbf{Distribution Similarity.} Following previous works~\cite{snapvideo,imagenvideo,latte}, we use (15) \textbf{\textit{FVD}}~\cite{fvd}, (16) \textbf{\textit{FID}}~\cite{fid}, (17) \textbf{\textit{KID}}~\cite{kid} to evaluate the distribution similarity of generated and training data.

\section{Experiments}
\label{sec:experiment}

\subsection{Model Design of MiraDiT}
\label{sec:model_design_of_miradit}

\begin{figure}[htbp]
\centering
  \includegraphics[width=1.0\linewidth,height=0.6\linewidth]{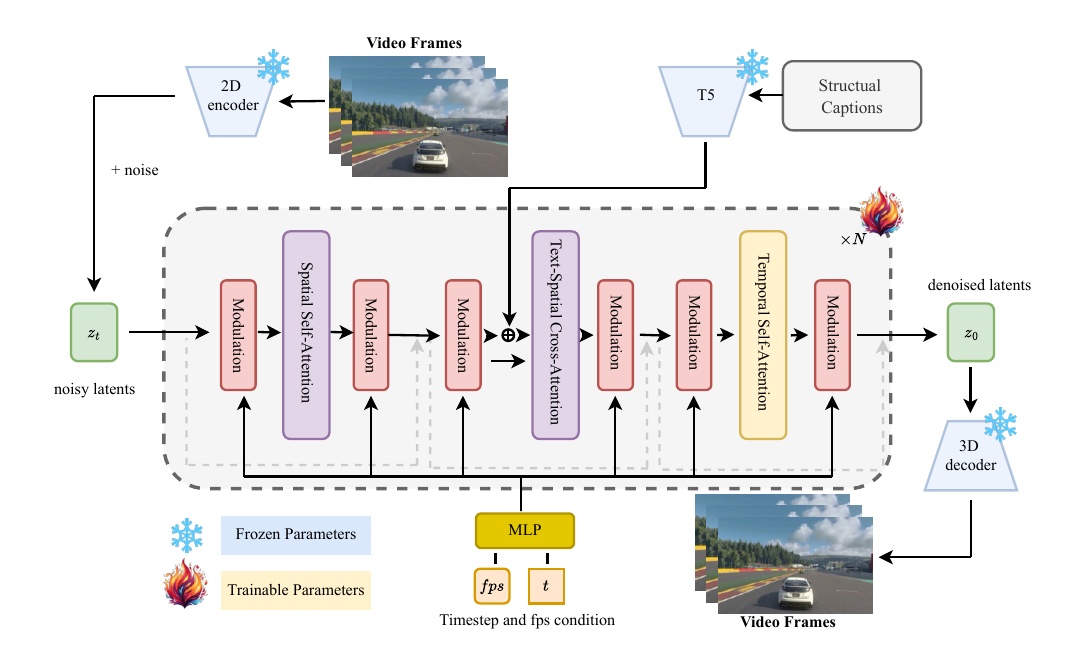}
\vspace{-4 mm}
\caption{\textbf{MiraDiT pipeline for long video generation.}}
\label{fig:miradit}
\vspace{-4 mm}
\end{figure}

To validate the effectiveness of MiraData for consistent long-video generation, we design an efficient pipeline based on Diffusion Transformer~\cite{dit}, as illustrated in Fig.\ref{fig:miradit}. Following SVD~\cite{svd}, we use a hybrid Variational Autoencoder with a 2D convolutional encoder and a 3D convolutional decoder to reduce flickering in generated videos.
Unlike previous methods\cite{svd, videocrafter1, dynamicrafter} that rely on short captions and typically use a CLIP text encoder with 77 output tokens, we employ a larger Flan-T5-XXL~\cite{flan-t5} for textual encoding, supporting up to 512 tokens for dense and structured caption understanding.

\paragraph{Text-spatial cross-attention.}
For latent denoising, we build a spatial-temporal transformer as the trainable generation backbone. 
As shown in Fig.\ref{fig:miradit}, we adopt spatial and temporal self-attention separately rather than full attention on all video pixels to reduce the heavy computational load of long-video generation. Similar to W.A.L.T~\cite{walt}, we apply extra conditioning on spatial queries during cross-attention to stabilize training and improve generation performance.
For faster convergence, we partially initialize spatial attention layers  from weights of text-to-image model Pixart-alpha~\cite{pixart}, while keeping other layers trained from scratch.  

\paragraph{FPS-conditioned modulation.}
Following DiT and Stable Diffusion 3~\cite{sd3}, we use a modulation mechanism for the current timestep condition. 
Additionally, we embed an extra current FPS condition in the AdaLN layer to enable motion strength control during inference in the generated videos.

\paragraph{Dynamic frame length and resolution.}
We train MiraDiT in a way that supports generating videos with different resolutions and lengths to evaluate the model performance on motion strength and 3D consistency in different scenarios. 
Inspired by NaViT~\cite{dehghani2024patch}, which uses Patch n' Pack to achieve dynamic resolution training, we apply a Frame n' Pack strategy to train videos with various temporal lengths. Specifically, we randomly drop frames with zero padding using a temporal mask, then apply masked self-attention and positional embeddings according to the temporal masks.
The gradients of masked frames are stopped as well.  
However, for varying resolution training, we didn't adopt Patch n' Pack since it made the model harder to train during our early experiments. Instead, we follow Pixart~\cite{pixart}  and use a bucket strategy where the models are trained on different resolution videos where each training batch only contains videos of the same resolution.

\paragraph{Inference details. }
During inference, we use the DDIM~\cite{ddim} sampler with 25 steps and classifier-free guidance of scale 12. The fps condition can be set between $5$ and $30$, allowing for flexibility in the generated video's frame rate. For evaluation purposes, we test all our models at 6 fps to ensure a consistent comparison across different settings.
To further enhance the visual quality of the generated videos, we provide an optional post-processing step using the RIFE~\cite{rife} model. By applying $4\times$ frame interpolation, we can increase the frame rate of the generated video to $24$ fps, resulting in smoother motion and improved overall appearance.

\vspace{-0.3cm}

\subsection{Comparison with Previous Video Generation Datasets}

\vspace{-0.3cm}

Our experiments aim to validate the effectiveness of MiraData in long video generation by assessing (1) temporal motion strength and consistency, and (2) visual quality and text alignment. We train MiraDiT models on WebVid-10M and MiraData separately, evaluating them on MiraBench at $384\times240$ resolution with 5s length using 14 metrics covering motion strength, consistency, visual quality, and text-video alignments.

Tab.~\ref{tab:compare_previous_dataset} shows that the model trained on MiraData demonstrates significant improvements in motion strength while maintaining temporal and 3D consistency compared to the WebVid-10M model. Moreover, MiraData's higher-quality videos and dense, accurate prompts lead to better visual quality and text-video alignments in the trained model.
We compare our MiraDiT model trained on MiraData to state-of-the-art open-source methods, OpenSora~\cite{opensora} (DiT-based) and VideoCrafter2~\cite{videocrafter2} (U-Net-based). Our model significantly outperforms previous methods in terms of motion strength and 3D consistency while achieving competitive results in visual quality and text-video alignment. This demonstrates MiraData's effectiveness in enhancing long video generation. Note that distribution-based metrics like FVD are not reported due to the difference in training datasets. More visual and metric comparisons are in the Appendix.

\begin{table}[htbp]
\small
    \caption{\textbf{Comparison of MiraDiT trained on MiraData and WebVid-10M~\cite{webvid10m}.} $\uparrow$ and $\downarrow$ means higher/lower is better. 1) - 14) indicates indices of metrics in MiraBench (Sec.~\ref{sec:mirabench}), where DD for Dynamic Degree, TS for Tracking Strength, DTC for DINO Temporal Consistency, CTC for CLIP Temporal Consistency, TMS for Temporal Motion Smoothness, MAE for Mean Absolute Error, RMSE for Root Mean Square Error, AQ for Aesthetic Quality, IQ for Imaging Quality, CA for Camera Alignment, MOA for Main Object Alignment, BA for Background Alignment, SA for Style Alignmnet, and OA for Overall Alignment. Best shown in \textbf{blod}, and second best shown in \underline{underlined}.}
    \label{tab:compare_webvid}
    \setlength\tabcolsep{1pt}
    \scalebox{0.87}{
\begin{tabular}{c|ccccccc}
\toprule
\toprule
\multirow{2}{*}{\textbf{Metrics}} & \multicolumn{2}{c|}{\textbf{Temporal Motion Strength}}                             & \multicolumn{3}{c|}{\textbf{Temporal Consistency}} & \multicolumn{2}{c}{\textbf{3D Consistency}} \\
  & 1) DD$_{\uparrow}$ &  \multicolumn{1}{c|}{2) TS$_\uparrow$}                      & 3) DTC$_\uparrow$ & 4) CTC$_\uparrow$      & \multicolumn{1}{c|}{5) TMS$_\uparrow$}   & 6) MAE$_{\downarrow \times 10^{-2}}$     & 7) RMSE$_{\downarrow \times 10^{-1}}$     \\ \midrule
OpenSora~\cite{opensora}               &     \underline{7.65}       &       \multicolumn{1}{c|}{16.07}                        &    12.34    &  13.20         & \multicolumn{1}{c|}{13.70}          &    \textbf{75.45}    &      \textbf{10.39}                        \\ 
VideoCrafter2~\cite{videocrafter2}               &     1.71       &       \multicolumn{1}{c|}{6.72}                        &    6.41    &  6.36        & \multicolumn{1}{c|}{6.60}          &   101.55    &     13.05                        \\ \midrule
MiraDiT (WebVid-10M~\cite{webvid10m})              &     7.12       &       \multicolumn{1}{c|}{\underline{22.36}}                        &    \underline{20.24}    &  \underline{20.97}         & \multicolumn{1}{c|}{21.86}          &    91.48     &      12.11                         \\
MiraDiT (\DatasetName)                &    \textbf{15.46}     &     \multicolumn{1}{c|}{\textbf{49.47}}                            &   \textbf{43.78}   &     \textbf{45.95}         & \multicolumn{1}{c|}{\textbf{47.24}}          &   \underline{85.27}      &       \underline{11.74}                         \\ 
\midrule \midrule
\multirow{2}{*}{\textbf{Metrics}} & \multicolumn{2}{c|}{\textbf{Visual Quality}}    & \multicolumn{5}{c}{\textbf{Text-Video Alignmnet}}  \\
                         & 8) AQ$_{\uparrow \times 10^{-}}$ & \multicolumn{1}{c|}{9) IQ$_\uparrow$} & 10) CA$_\uparrow$                      & 11) MOA$_\uparrow$     & 12) BA$_\uparrow$                        & 13) SA$_\uparrow$ & 14) OA$_\uparrow$    \\ \midrule
OpenSora~\cite{opensora}               &    47.10       &       \multicolumn{1}{c|}{59.54}                        &    \underline{12.40}   &  \textbf{18.12 }        & \textbf{13.20}         &   \textbf{13.35}    &    16.12                   \\ 
VideoCrafter2~\cite{videocrafter2}               &    \textbf{58.69}       &       \multicolumn{1}{c|}{\textbf{64.96}}                        &   12.00  &  \underline{17.90}        & 11.25        &  12.15    &       \textbf{16.90}       \\ \midrule
MiraDiT (WebVid-10M~\cite{webvid10m})                &    43.11     & \multicolumn{1}{c|}{58.58}        &       12.35                       &          14.32    &      11.90                          &    12.32     &      15.31           \\
MiraDiT   (\DatasetName)                &    \underline{49.90}     & \multicolumn{1}{c|}{\underline{63.71}}        &             \textbf{12.66}                 &        14.67      &    \underline{12.18}                            &    \underline{12.59}     &   \underline{ 16.66}           \\ \bottomrule \bottomrule
\end{tabular}
}
\vspace{-10pt}
\end{table}

\subsection{Role of Caption Length and Granularity}
We investigate the impact of caption length and granularity on MiraDiT's performance by evaluating the model using short, dense, and structural captions separately. The results in Tab.~\ref{tab:ablation_caption} demonstrate that longer and more detailed captions do not necessarily improve the visual quality of the generated videos. However, they offer significant benefits in terms of increased dynamics, enhanced temporal consistency, more accurate generation control, and better alignment between the text and the generated video content. These findings highlight the importance of caption granularity in guiding the model to produce videos that more closely match the desired descriptions while maintaining coherence and realism.
Please see appendix for more qualitative results and detailed ablation studies.

\begin{table}[htbp]
\vspace{-4 mm}
\small
\centering
\caption{ \textbf{Comparison of MiraDiT model with different caption length and granularity.}  1) - 14) indicates indices of metrics in MiraBench (Sec.~\ref{sec:mirabench}). See Tab.~\ref{tab:compare_webvid} for the meaning of metrics annotation.}
    \label{tab:ablation_caption}
    \setlength\tabcolsep{3.5pt}
\begin{tabular}{c|cccccccc}
\toprule
\textbf{Metrics} & 1) DD$_{\uparrow}$ &  \multicolumn{1}{c|}{ 2) TS$_\uparrow$}                      & 3) DTC$_\uparrow$ &  4) CTC$_\uparrow$      &  \multicolumn{1}{c|}{5) TMS$_\uparrow$}   &   8) AQ$_{\uparrow}$ & \multicolumn{1}{c|}{9) IQ$_\uparrow$} & 14) OA$_\uparrow$     \\ \midrule
Short Caption          &     9.45       &       \multicolumn{1}{c|}{27.03}                        &    24.39    &  25.20        & \multicolumn{1}{c|}{26.05}          &   4.84    &    \multicolumn{1}{c|}{63.64}   & 7.73\\ 
Dense Caption         &     17.39      &       \multicolumn{1}{c|}{52.53}                        &  46.13   &  48.35       & \multicolumn{1}{c|}{50.12}          &   5.14   &      \multicolumn{1}{c|}{63.43} &      14.88                  \\ 
Structural Caption            &     19.53      &       \multicolumn{1}{c|}{68.85}                        &  60.83   & 64.31         & \multicolumn{1}{c|}{65.56}          &  4.99   &      \multicolumn{1}{c|}{64.07}      & 15.36                \\ \bottomrule 
\end{tabular}
\vspace{-10pt}
\end{table}

\section{Conclusion and Discussion}
\label{sec:conclusion}

\vspace{-0.3cm}

\textbf{Conclusion.} In conclusion, \DatasetName~complements existing video datasets with high-quality, long-duration videos featuring detailed captions and strong motion intensity. 
Curated from diverse video sources and annotated with multiple high-performance models, \DatasetName~shows advantages in comprehensive evaluation framework \BenchName~with the designed \ModelName~model, highlighting its potential to push the boundaries of high-motion, temporally consistent long video generation.

\textbf{Limitation.} Despite \DatasetName's advantages over previous datasets, it still has limitations, such as inherent biases, potential annotation errors, and insufficient coverage. The evaluation metrics in \BenchName~may also yield inaccurate results in uncommon video scenarios, such as jitter or overexposure. Due to the page limit, the appendix will provide a detailed discussion.

\textbf{Potential Negative Societal Impacts.} The enhanced video generation capabilities promoted by \DatasetName~could lead to negative societal impacts and ethical issues, including the creation of deepfakes and misinformation, privacy breaches, and harmful content generation. We would engage in implementing stringent ethical guidelines, ensuring robust privacy protections, and promoting unbiased dataset curation to prevent these issues. The appendix provides a detailed discussion.

\clearpage

{
\bibliographystyle{ieeetr}
\bibliography{reference}
}

\end{document}


\maketitle

\appendix

\section{Appendix}

Include extra information in the appendix. This section will often be part of the supplemental material. Please see the call on the NeurIPS website for links to additional guides on dataset publication.

\begin{enumerate}

\item Submission introducing new datasets must include the following in the supplementary materials:
\begin{enumerate}
  \item Dataset documentation and intended uses. Recommended documentation frameworks include datasheets for datasets, dataset nutrition labels, data statements for NLP, and accountability frameworks.
  \item URL to website/platform where the dataset/benchmark can be viewed and downloaded by the reviewers.
  \item URL to Croissant metadata record documenting the dataset/benchmark available for viewing and downloading by the reviewers. You can create your Croissant metadata using e.g. the Python library available here: https://github.com/mlcommons/croissant
  \item Author statement that they bear all responsibility in case of violation of rights, etc., and confirmation of the data license.
  \item Hosting, licensing, and maintenance plan. The choice of hosting platform is yours, as long as you ensure access to the data (possibly through a curated interface) and will provide the necessary maintenance.
\end{enumerate}

\item To ensure accessibility, the supplementary materials for datasets must include the following:
\begin{enumerate}
  \item Links to access the dataset and its metadata. This can be hidden upon submission if the dataset is not yet publicly available but must be added in the camera-ready version. In select cases, e.g when the data can only be released at a later date, this can be added afterward. Simulation environments should link to (open source) code repositories.
  \item The dataset itself should ideally use an open and widely used data format. Provide a detailed explanation on how the dataset can be read. For simulation environments, use existing frameworks or explain how they can be used.
  \item Long-term preservation: It must be clear that the dataset will be available for a long time, either by uploading to a data repository or by explaining how the authors themselves will ensure this.
  \item Explicit license: Authors must choose a license, ideally a CC license for datasets, or an open source license for code (e.g. RL environments).
  \item Add structured metadata to a dataset's meta-data page using Web standards (like schema.org and DCAT): This allows it to be discovered and organized by anyone. If you use an existing data repository, this is often done automatically.
  \item Highly recommended: a persistent dereferenceable identifier (e.g. a DOI minted by a data repository or a prefix on identifiers.org) for datasets, or a code repository (e.g. GitHub, GitLab,...) for code. If this is not possible or useful, please explain why.
\end{enumerate}

\item For benchmarks, the supplementary materials must ensure that all results are easily reproducible. Where possible, use a reproducibility framework such as the ML reproducibility checklist, or otherwise guarantee that all results can be easily reproduced, i.e. all necessary datasets, code, and evaluation procedures must be accessible and documented.

\item For papers introducing best practices in creating or curating datasets and benchmarks, the above supplementary materials are not required.
\end{enumerate}

Misinformation and Deepfakes: Enhanced video generation capabilities can be misused to create highly realistic fake videos, potentially spreading misinformation or propaganda. This can undermine trust in digital media and have serious implications for public discourse, elections, and personal reputations.
Privacy Concerns: The creation and dissemination of realistic synthetic videos raise significant privacy issues. Individuals may be depicted in situations they were never a part of, leading to potential defamation or unauthorized use of likenesses.
Content Moderation Challenges: With the ability to generate high-quality, realistic videos comes the challenge of moderating and controlling the spread of harmful content. Platforms may struggle to effectively identify and mitigate the distribution of violent, explicit, or otherwise harmful generated videos.
Intellectual Property Violations: The use of curated video sources and the generation of new content can lead to intellectual property concerns. There is a risk of infringing on copyrights, trademarks, or other intellectual property rights, leading to legal disputes and ethical challenges.
Bias and Representation: The dataset curation process, despite efforts to be diverse, might still reflect inherent biases. These biases can propagate into generated content, reinforcing stereotypes or marginalizing certain groups. Ensuring fair and unbiased representation in generated videos remains a complex challenge.
Economic Displacement: Advances in video generation technology threaten jobs in industries reliant on human creativity and labor, such as film production, video editing, and content creation. This could lead to economic displacement and require significant workforce retraining initiatives.
Psychological Impact: The proliferation of highly realistic synthetic videos can have psychological impacts on viewers. It can lead to confusion, anxiety, or distress when distinguishing between real and fake content becomes increasingly difficult.
Ethical Use of AI: The powerful capabilities of AI-driven video generation necessitate ongoing ethical considerations. Ensuring that these tools are used responsibly and in ways that benefit society, rather than harm it, requires robust guidelines, regulations, and ethical frameworks.

(1) Scope of Sources: Despite our efforts to include diverse sources, there is an inherent bias in any curated dataset. Certain video genres or content types may be underrepresented, potentially limiting the generalizability of models trained on \DatasetName. (2) Annotation Quality: Although we utilize a well-designed process for video selection and use GPT-4V to generate structured captions and detailed descriptions, automatic annotation processes can introduce errors or inconsistencies. Manual verification, while extensive, cannot guarantee the complete accuracy and semantic consistency of all annotations. (3) Video Length: While \DatasetName includes longer videos than previous datasets, the average duration may still not fully capture the complexity and nuances of real-world video content, especially for applications requiring extended temporal coherence. (4) Evaluation Metrics: Although \BenchName introduces novel metrics for assessing temporal consistency and motion strength, these metrics are still subject to ongoing validation. (5) Model Dependency: The experiments conducted to demonstrate the utility of \DatasetName are based on our DiT-based video generation model, \ModelName. The observed advantages may not directly translate to other models or architectures, necessitating additional validation across diverse generative frameworks. (6) Ethical Considerations: The manual selection and curation process for \DatasetName, while aiming for high quality, might inadvertently include content with ethical or copyright concerns. Continuous monitoring and ethical review are necessary to ensure compliance with content usage guidelines. (7) Generalizability: The specific enhancements made in \DatasetName, such as increased motion intensity and detailed captions, may be particularly suited for certain applications but not for others. Researchers must consider the domain-specific applicability of \DatasetName when utilizing it for their projects.

By acknowledging these limitations, we aim to provide a balanced perspective on the contributions of \DatasetName and encourage further research to address these challenges.

While MiraData and the associated advancements in video generation offer significant benefits, it is important to consider potential negative societal impacts:

Misinformation and Deepfakes: Enhanced video generation capabilities can be misused to create highly realistic fake videos, potentially spreading misinformation or propaganda. This can undermine trust in digital media and have serious implications for public discourse, elections, and personal reputations.
Privacy Concerns: The creation and dissemination of realistic synthetic videos raise significant privacy issues. Individuals may be depicted in situations they were never a part of, leading to potential defamation or unauthorized use of likenesses.
Content Moderation Challenges: With the ability to generate high-quality, realistic videos comes the challenge of moderating and controlling the spread of harmful content. Platforms may struggle to effectively identify and mitigate the distribution of violent, explicit, or otherwise harmful generated videos.
Intellectual Property Violations: The use of curated video sources and the generation of new content can lead to intellectual property concerns. There is a risk of infringing on copyrights, trademarks, or other intellectual property rights, leading to legal disputes and ethical challenges.
Bias and Representation: The dataset curation process, despite efforts to be diverse, might still reflect inherent biases. These biases can propagate into generated content, reinforcing stereotypes or marginalizing certain groups. Ensuring fair and unbiased representation in generated videos remains a complex challenge.
Economic Displacement: Advances in video generation technology threaten jobs in industries reliant on human creativity and labor, such as film production, video editing, and content creation. This could lead to economic displacement and require significant workforce retraining initiatives.
Psychological Impact: The proliferation of highly realistic synthetic videos can have psychological impacts on viewers. It can lead to confusion, anxiety, or distress when distinguishing between real and fake content becomes increasingly difficult.
Ethical Use of AI: The powerful capabilities of AI-driven video generation necessitate ongoing ethical considerations. Ensuring that these tools are used responsibly and in ways that benefit society, rather than harm it, requires robust guidelines, regulations, and ethical frameworks.